\documentclass[11pt]{article}

\usepackage[final]{acl}

\usepackage{times}
\usepackage{latexsym}
\usepackage[T1]{fontenc}
\usepackage[utf8]{inputenc}
\usepackage{microtype}
\usepackage{inconsolata}
\usepackage{graphicx}
\usepackage{booktabs}
\usepackage{amsmath}

\title{The Curse of Multilinguality in Lexical Normalization}

\author{Saman Rahbar \\
  Independent Researcher \\
  \texttt{info@srahbar.com}}

\begin{document}
\maketitle

\begin{abstract}
Lexical normalization rewrites the noisy, non-standard words that fill
user-generated text (\emph{tmrw}, \emph{u}, \emph{gr8}) into their standard forms. Because labelled data is scarce for most languages, a popular shortcut
is to train a single model on many languages at once. We ask a simple
question: how many languages should such a model be trained on? Using one
fixed-capacity character-level model and twelve languages from a standard
benchmark, we vary the number of jointly trained languages from one to twelve
and measure per-language accuracy. We find a clear curse of multilinguality:
accuracy is highest when a language is trained with only a few others, often just one to four, and then falls steadily and substantially, dropping by about
forty percent as the rest are piled on. A control that holds the total amount of
training data constant makes the decline arrive sooner and fall further, which
points to competition among the languages for one fixed-size model rather than
to how much data is available. We also test whether a language's typological
distance from the others predicts its ideal number of co-training languages, and
find no dependable rule: any apparent relationship rests on a couple of languages
and does not hold up. For compact normalization models, less can be more: a few
languages beat pooling everything into a single model.
\end{abstract}

\section{Introduction}

Text written by people online (social-media posts, comments, messages) is
full of abbreviations, misspellings, phonetic spellings, and playful
respellings. \emph{Lexical normalization} is the task of turning each
such non-standard word into its canonical form, mapping \emph{tmrw} to
\emph{tomorrow} or \emph{u} to \emph{you}, so that downstream tools built for
clean text can cope. It is a long-standing preprocessing step for noisy-text
natural language processing (NLP) \cite{han2011lexical}, and a shared task in
its own right \cite{baldwin2015}.

The main obstacle is data. Hand-labelled normalization corpora are small and
exist for only a handful of languages. The usual way around this is to train one
model on several languages at once, in the hope that they reinforce one another:
a misspelling pattern learned in one language may transfer to another, and the
pooled data is larger than any single language's. The \textsc{MultiLexNorm}
shared task was organized around exactly this setting
\cite{vandergoot2021multilexnorm}, and its strongest entry was a single
multilingual byte-level model fine-tuned on all languages together
\cite{samuel2021ufal}; the same pool-everything recipe underpins multilingual
pretraining more broadly \cite{conneau2020unsupervised}.

But is more always better? In machine translation and in large language
models, researchers have repeatedly observed a \emph{curse of multilinguality}
\cite{conneau2020unsupervised,chang2024curse}: when a model of fixed size is
trained on more and more languages, each individual language eventually gets
\emph{worse}, because the languages compete for the model's limited capacity.
Whether the same trade-off governs lexical normalization has not, to our
knowledge, been studied. Normalization is a small, character-level task with
tiny models and short outputs, so it is not obvious that the crowded-capacity story carries over: the models may be far from their capacity limit, or the
shared character patterns across languages may make extra languages purely
helpful.

We study this directly. We fix one small character-level model and, holding its
size constant, train it on every number of languages from one to twelve, drawn
from the \textsc{MultiLexNorm} benchmark, and measure how well it normalizes each
individual language. Our contributions are:

\begin{itemize}
  \item We map the accuracy-versus-number-of-languages frontier for lexical
  normalization and show that per-language accuracy peaks at a few languages and
  then declines steadily, and the curse of multilinguality holds
  (Section~\ref{sec:frontier}).
  \item Through a control that holds total training data fixed, we show the
  decline is not explained by data volume but by many languages sharing one
  fixed-size model (Section~\ref{sec:ablation}).
  \item We test, and fail to confirm, the intuitive idea that a language's
  typological isolation predicts its ideal number of co-training languages: any
  apparent effect is fragile and rests on a couple of languages
  (Section~\ref{sec:typology}).
\end{itemize}

\section{Related Work}

\paragraph{Lexical normalization.}
Normalization of user-generated text goes back to work on text messages and
tweets \cite{han2011lexical} and was shaped by the W-NUT shared tasks
\cite{baldwin2015} and by broad-coverage systems such as \textsc{MoNoise}
\cite{vandergoot2017monoise}. We build directly on \textsc{MultiLexNorm}
\cite{vandergoot2021multilexnorm}, which put twelve normalization datasets into
one word-aligned format under one metric and made cross-language comparison
possible for the first time. Its strongest entry fine-tuned the byte-level model
\textsc{ByT5} \cite{samuel2021ufal,xue2022byt5} on all languages jointly. That
is the setting we take apart: the benchmark fixed the languages and the metric,
but the number of languages was never treated as a variable.

\paragraph{The curse of multilinguality.}
Adding languages to a fixed-capacity model helps for a while and then hurts each
one. This has been documented for massively multilingual translation
\cite{arivazhagan2019massively}, for cross-lingual language models
\cite{conneau2019xlm,conneau2020unsupervised}, and most directly in a study of
language-model perplexity spanning 250 languages \cite{chang2024curse}. The
mechanism usually named is negative interference, where languages start to
degrade one another once too little capacity is left to go around
\cite{wang2020negative}, and the usual remedy is to give each language a little
private capacity, as in adapter modules \cite{pfeiffer2020madx}. Every one of
these results comes from large models doing word- or subword-level work.
Normalization is neither: the models are tiny, the alphabet is shared, and the
output is a handful of characters. Whether a curse appears at that scale is an
open question, and a plausible case can be made either way, since a small model
might be nowhere near its capacity limit, or shared character patterns might make
every extra language a gift.

\paragraph{Typological distance.}
If languages do compete, one might expect similar ones to compete less. Distance
between languages can be read off \textsc{URIEL} and its \texttt{lang2vec}
interface \cite{littell2017uriel}, which we use to ask whether a language's
typological isolation says anything about how many partners it wants.

\section{Experimental Setup}
\label{sec:setup}

\paragraph{Data and metric.}
We use all twelve language datasets from \textsc{MultiLexNorm}: Danish, German,
English, Spanish, Croatian, Indonesian--English, Italian, Dutch, Slovenian,
Serbian, Turkish, and Turkish--German (two are code-switched pairs). The data is
pre-tokenized and word-aligned: each input token is paired with its gold
normalized form, and most tokens are already standard and map to themselves.
Table~\ref{tab:data} lists the sizes; the datasets differ widely both in size
(from about 6k to 57k training tokens) and in how much normalization they need
(from $6.6\%$ to $37\%$ of tokens changed), which matters for how much a language
can gain from partners. We keep the official train/test splits so that our per-language numbers sit on
the same footing as the published \textsc{MultiLexNorm} results, and for the
five languages without a development split we hold out $10\%$ of training as
development. No data outside these training splits is used at any point: the
model has no pretraining stage, and every run below starts from random
initialization.

We report the benchmark's official \emph{Error Reduction Rate} (ERR), which
measures how much of the achievable normalization a system performs relative to
doing nothing:
\begin{equation}
\text{ERR} = \frac{\text{acc} - \text{LAI}}{1 - \text{LAI}},
\end{equation}
where \text{acc} is token accuracy and \text{LAI} (``leave-as-is'') is the
accuracy of copying the input unchanged. ERR${}=1$ is perfect; ERR${}=0$ means
no better than copying; ERR${}<0$ means the system corrupts more than it fixes.

\begin{table}[t]
\centering\small
\begin{tabular}{lrrr}
\toprule
Lang & Train & Test & \% changed \\
\midrule
Danish        & 16{,}448 & 3{,}758  & 9.2 \\
German        & 15{,}006 & 5{,}082  & 17.2 \\
English       & 35{,}216 & 29{,}421 & 7.6 \\
Spanish       & 7{,}189  & 6{,}635  & 7.7 \\
Croatian      & 54{,}416 & 15{,}695 & 6.6 \\
Indo.--Eng.   & 13{,}949 & 4{,}366  & 14.7 \\
Italian       & 12{,}645 & 1{,}996  & 7.3 \\
Dutch         & 12{,}381 & 5{,}413  & 29.7 \\
Slovenian     & 44{,}944 & 15{,}021 & 14.8 \\
Serbian       & 56{,}823 & 17{,}119 & 7.4 \\
Turkish       & 6{,}443  & 1{,}639  & 37.0 \\
Turkish--Ger. & 12{,}773 & 3{,}735  & 24.1 \\
\bottomrule
\end{tabular}
\caption{The twelve \textsc{MultiLexNorm} datasets: training and test tokens,
and the share of training tokens whose gold form differs from the input
(``\% changed'').}
\label{tab:data}
\end{table}

\paragraph{Model.}
Every experiment uses the \emph{same} model: a character-level
encoder--decoder Transformer \cite{vaswani2017attention} that reads a raw token
one character at a time and generates its normalized form one character at a
time. Encoder and decoder each have three layers, a model width of $128$, four
attention heads, and a feed-forward width of $512$; a single character
vocabulary (the union of characters over all languages, plus start, end, and
padding symbols) is shared by every language. This comes to $1.49$M parameters
in total (Table~\ref{tab:hyper}). Holding this model fixed across every setting
is the crux of the study: any change in per-language accuracy as we add
languages reflects how that fixed capacity is shared, not a larger or smaller
model. We keep the model small and train it from scratch, rather than fine-tuning a
pretrained byte-level model such as \textsc{ByT5} \cite{xue2022byt5}, for one
reason: a pretrained model arrives with an unknown amount of multilingual
capacity already inside it, and we would then be measuring how that inherited
capacity gets redistributed, not how a known, fixed capacity is shared.
Training from scratch makes capacity a quantity we set, not one we inherit. The cost of that choice is absolute performance, and we return to it
when two languages turn out to be normalized worse than not at all.

\begin{table}[t]
\centering\small
\begin{tabular}{ll}
\toprule
Setting & Value \\
\midrule
Architecture      & char Transformer (enc--dec) \\
Layers (enc / dec) & 3 / 3 \\
Model width        & 128 \\
Attention heads    & 4 \\
Feed-forward width & 512 \\
Parameters         & 1.49M \\
Max token length   & 32 characters \\
Optimizer          & AdamW, lr $3\!\times\!10^{-4}$ \\
Batch size         & 256 tokens \\
Max steps          & 8{,}000 (early stopping) \\
Seeds              & 3 \\
\bottomrule
\end{tabular}
\caption{Model and training settings, identical across all conditions.}
\label{tab:hyper}
\end{table}

\paragraph{Protocol.}
For each number of languages $k$ from $1$ to $12$, we train the model jointly on
random size-$k$ subsets of the twelve languages. Subsets are drawn with a
coverage constraint so that every language appears in a comparable number of them; otherwise, at small $k$, some languages would never be trained. We repeat
the whole sweep over three random seeds, giving $390$ training runs.
Training pools the chosen languages' data, shuffles it, and optimizes with
AdamW; we stop early when mean development ERR over the trained languages stops
improving, and keep the best checkpoint. We then measure each trained model's
ERR on the test set of every language it was trained on. For a language, its
mean ERR across all subsets of a given size, plotted against $k$, is its
\emph{frontier}, and the $k$ at which that mean is highest is its optimum
$k^\ast$. Because greedy decoding of full test sets dominates runtime, during
the sweep we estimate each language's test ERR on a fixed sample of up to
$2{,}000$ test tokens.

\paragraph{Data-controlled ablation.}
Adding languages does two things at once: it changes the mix the model must
serve, and it increases the total training data. To separate them, we repeat
the entire protocol while capping the total number of training tokens at a
fixed budget, so that a twelve-language model sees no more data than a
one-language model. Any decline that survives this control cannot be a
data-volume effect.

\paragraph{Typology.}
For each language we compute its \emph{isolation}: the mean \textsc{URIEL}
syntactic distance to the other eleven. Turkish and Indonesian--English are the
two clear outliers; the European languages cluster tightly.

\section{Results}

\subsection{Normalization has a curse of multilinguality}
\label{sec:frontier}

\begin{figure}[t]
\centering
\includegraphics[width=\columnwidth]{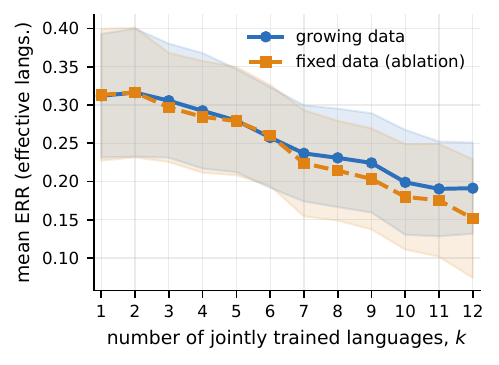}
\caption{Mean ERR against the number of jointly trained languages $k$, averaged
over the ten languages the model normalizes effectively. Both settings peak at $k=2$ and then decline steadily: the curse of multilinguality. Holding total
data fixed (dashed) makes the drop steeper, so the decline is not a matter of
data volume. Shaded bands are 95\% confidence intervals across languages.}
\label{fig:frontier}
\end{figure}

Figure~\ref{fig:frontier} shows the frontier. Two languages, Spanish and
Italian, are set aside first: on both the model scores below zero at every $k$,
meaning it corrupts more than it repairs, so a best number of partners is not
defined for them. We take them up in their own right below, and average here
over the remaining ten.

Across those ten, mean ERR peaks at $k=2$ ($0.316$), barely above training alone
($0.312$), and then falls steadily to $0.191$ at $k=12$, about $40\%$ below the
peak. The fall is not an artifact of averaging noisy curves. Every one of the ten
languages ends lower at $k=12$ than at its own best, and nine of the ten are
worse with eleven partners than with none. Six peak at an interior $k^\ast$; the
remaining four are best trained alone, and the mean optimum across all ten is
$k^\ast=2.5$. A few partners can help, but most only crowd.

\begin{figure}[t]
\centering
\includegraphics[width=\columnwidth]{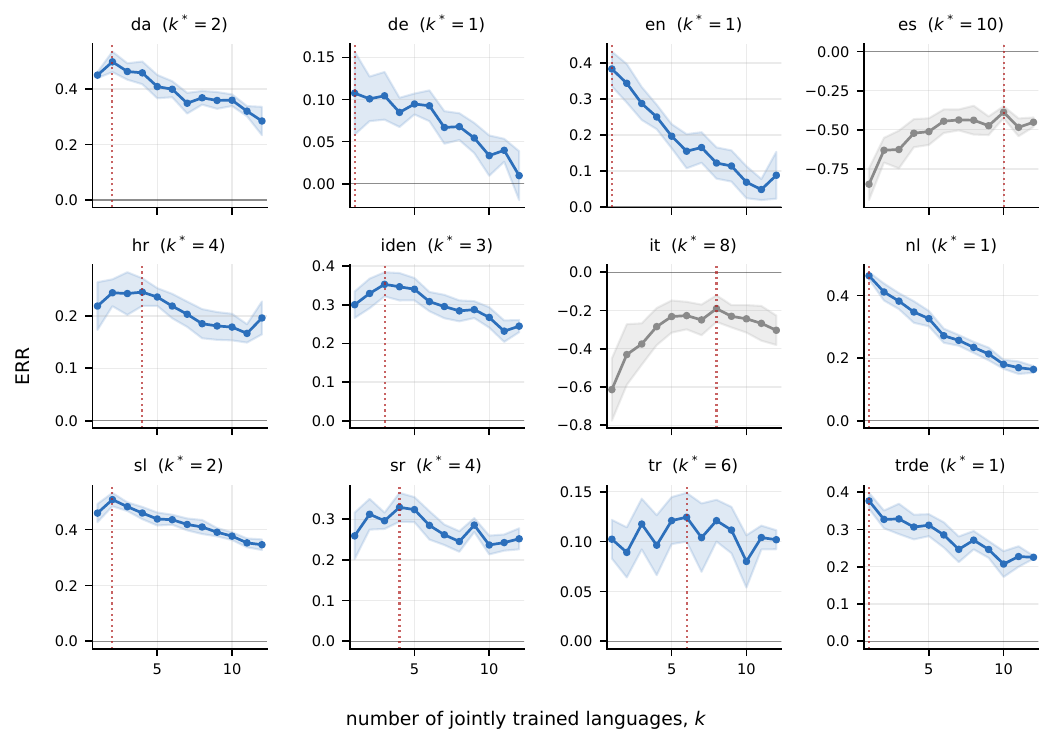}
\caption{Per-language frontiers (growing-data setting). Dotted lines mark each
language's optimum $k^\ast$. Languages well served alone (e.g.\ English, Dutch) gain nothing from partners; the rest benefit from a few before crowding hurts.}
\label{fig:perlang}
\end{figure}

The per-language curves (Figure~\ref{fig:perlang}, Table~\ref{tab:perlang}) fill
in the picture. English is the sharpest case: it does best on its own, at an ERR of $0.384$, and the trend runs downward from there to $0.088$ at $k=12$. English has enough
data to stand alone, so sharing capacity only dilutes it. The four languages
whose optimum is $k^\ast=1$ (English, German, Dutch, Turkish--German) gain
nothing from any partner. The other six instead take a few
partners before crowding sets in: Danish and Slovenian peak at $k=2$,
Indonesian--English at $k=3$, Croatian and Serbian at $k=4$, and Turkish, the
smallest dataset in the benchmark and the one needing the heaviest normalization,
at $k=6$, later than any other language among the ten, though its ERR reaches
only $0.125$ even at its best.

\paragraph{Why Spanish and Italian fail, and what follows for the claim.}
These two behave unlike the rest, and the pattern is worth stating exactly,
because it bears on how the aggregate should be read. Both score below zero at
every $k$: the model changes more tokens than it repairs. Both are at their
\emph{worst} trained alone ($-0.847$ Spanish, $-0.614$ Italian), both are better
at twelve languages than at one ($-0.451$ and $-0.304$), and both reach their
best at $k=10$ and $k=8$, later than any other language here, where the largest
optimum among the remaining ten is $k^\ast=6$.

That direction rules out one explanation. Were the failure caused by pooling
with languages that normalize more heavily, adding partners would deepen it;
instead adding partners reduces it, and the single-language setting, in which no
partner exists, is the worst case of all. What the direction is consistent with
is a shortage of in-language signal. Spanish and Italian pair a low change rate with a
small training set: $7.7\%$ and $7.3\%$ of tokens require any change, against
$37.0\%$ for Turkish, on only $7{,}189$ and $12{,}645$ training tokens
(Table~\ref{tab:data}). That combination leaves them the fewest changed tokens of
any dataset here, roughly $550$ for Spanish and $920$ for Italian, where no other
language falls below $1{,}500$. Croatian has an even lower change rate ($6.6\%$)
but a much larger training set, so it sees about $3{,}600$ changed tokens and
does not fail. What is scarce for Spanish and Italian is the absolute amount of
in-language evidence about what to change, not the proportion.
We did not analyse the model's error types, so we give this as the reading the
data supports and not as a measured mechanism.

One consequence is definitional. Their $k^\ast$ values are not optima in the
sense the other ten have; they mark the least-harm point on a curve that never
becomes useful, which is why a best number of partners is not a meaningful
quantity for them.

Two consequences bear on the headline. First, these two serve as an internal
control: the same sweep, applied where the binding constraint differs, yields a
rising curve instead of a falling one, so the decline in the other ten is not an
artifact of how subsets were drawn or how ERR was computed. Second, the aggregate
does not depend on excluding them. Averaging all twelve still gives a clear
curse, peaking at $0.177$ ($k=4$) and falling to $0.097$ at $k=12$, a $45\%$ drop
against the $40\%$ we report for the ten. We exclude them to avoid averaging over
curves whose shape reflects a floor effect, not in order to obtain the result.

What this does bound is scope. At the capacity we fixed the model is not merely
crowded; it is below the level needed to serve the two datasets that carry the
fewest changed tokens at all.
The frontier we map is therefore the frontier for a model of that size, and a
larger one might normalize both successfully and peak later.

\begin{table}[t]
\centering\small
\begin{tabular}{lrrrr}
\toprule
Lang & $k^\ast$ & ERR$_{k=1}$ & ERR$_{k^\ast}$ & ERR$_{k=12}$ \\
\midrule
Danish        & 2  & .450 & .498 & .285 \\
German        & 1  & .108 & .108 & .010 \\
English       & 1  & .384 & .384 & .088 \\
Spanish       & 10 & $-$.847 & $-$.387 & $-$.451 \\
Croatian      & 4  & .219 & .246 & .196 \\
Indo.--Eng.   & 3  & .300 & .353 & .245 \\
Italian       & 8  & $-$.614 & $-$.190 & $-$.304 \\
Dutch         & 1  & .464 & .464 & .164 \\
Slovenian     & 2  & .459 & .507 & .346 \\
Serbian       & 4  & .259 & .329 & .252 \\
Turkish       & 6  & .102 & .125 & .102 \\
Turkish--Ger. & 1  & .377 & .377 & .225 \\
\bottomrule
\end{tabular}
\caption{Per-language results (growing-data setting). $k^\ast$ is the optimal
number of jointly trained languages; ERR is shown at one language, at the
optimum, and at all twelve. Spanish and Italian have negative ERR throughout and, unlike every other
language, improve as partners are added; they are excluded from the aggregate
frontier and discussed in Section~\ref{sec:frontier}. Among the other ten, six
peak at an interior $k^\ast$.}
\label{tab:perlang}
\end{table}

\subsection{The decline is not a data-volume effect}
\label{sec:ablation}

A natural worry is that the drop simply reflects data volume: in our main
setting the total training pool grows with $k$, and one might suspect the
frontier tracks that. The fixed-data-budget control (dashed line in
Figure~\ref{fig:frontier}) rules this out. With the total number of training
tokens capped, the peak and the decline remain; in fact the fall is
\emph{steeper} (about $52\%$ at $k=12$ versus $40\%$) and the per-language
optima move earlier (mean $k^\ast$ from $2.5$ to $2.1$). Whether we let the data
grow or hold it fixed, adding languages past a small number hurts.

The decline is therefore a cost of making one fixed-size model serve more
languages at once, not a matter of how much data is available. Several mechanisms could contribute to that cost (fewer effective parameters per language, less training exposure per language under a fixed budget, and interference between languages' representations), and our design does not separate them; we refer to them together as \emph{capacity}. What the ablation
establishes is the negative result that matters in practice: the drop is not
bought back by adding data.

\subsection{Typology does not predict the sweet spot}
\label{sec:typology}

\begin{figure}[t]
\centering
\includegraphics[width=\columnwidth]{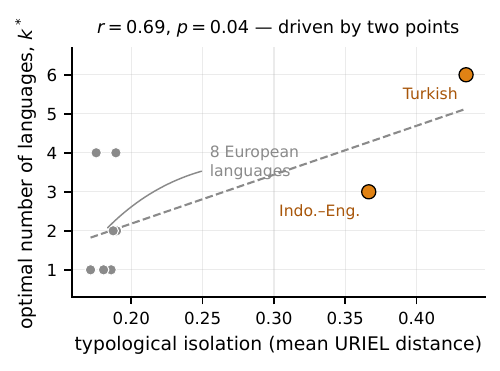}
\caption{Typological isolation versus optimum $k^\ast$ for the ten effective
languages. The positive correlation ($r=0.69$, $p=0.04$) rests entirely on the
two most isolated languages (Turkish, Indonesian--English); the European
languages cluster with no internal trend, and the correlation vanishes over all
twelve ($r=0.25$). We draw no reliable typological rule.}
\label{fig:typology}
\end{figure}

It is tempting to think a language that is typologically far from the others
should prefer fewer co-training partners. We find no dependable support, and if
anything the trend runs the other way. Correlating each effective language's
isolation with its optimum $k^\ast$ (Figure~\ref{fig:typology}) gives a positive coefficient ($r=0.69$, $p=0.04$), with isolated languages preferring \emph{more} partners, not fewer. But the figure shows why this should not be trusted: the
relationship rests entirely on the two most isolated languages, Turkish and
Indonesian--English, while the eight European languages form a tight cluster
with no internal trend. And it is brittle: including the two languages the
model fails on erases it ($r=0.25$, $p=0.45$ over all twelve). We therefore draw
no typological rule for how many languages to train on. What comes closer in our
data is how well a language is served on its own, with those that already do well
alone wanting fewer partners ($r=-0.58$), though across ten languages this too
falls short of significance ($p=0.08$) and we report it as the more promising
direction rather than a rule. Training-set size by itself predicts nothing
($r=0.18$, $p=0.62$).

\section{Discussion}

For anyone building a compact multilingual normalizer, the takeaway is concrete:
more languages are not automatically better. For most languages here the best model used only a few, typically one to four, and some languages were best off training alone. Pooling all twelve was, on average, worse than
using fewer.

The pattern looks like a trade-off. A few extra languages can help: five of the ten do best with one to three partners, and a sixth, Turkish, with five; plausibly some noise patterns
carry over between languages. Past that, adding languages hurts, and the
fixed-data control shows this is not bought back by more data: one fixed-size
model can only serve so many languages at once. The obvious lever is capacity. A
larger model should tolerate more languages before the curse sets in, and
remedies from large multilingual models, such as per-language adapter modules
that avoid sharing all parameters \cite{pfeiffer2020madx}, may transfer to this
setting. How the peak moves with model size, and whether lightweight per-language
capacity flattens it, is a natural next step.

Two limits should be read alongside the result rather than after it. We vary the
number of languages at one model size and one model family, so what we establish
is that a curse exists in this regime and where its peak falls here, not that the
peak sits at one to four languages generally. The Spanish and Italian failures
make the point concrete: at this capacity the model is not only crowded but below
the level needed to serve languages with little changed-token evidence at all,
and a model large enough to
fix that might well peak later. We would expect the shape to persist and the peak
to move right, but we have not measured it, and a reader planning a system should
treat ``a few languages'' as a finding about compact models, not a universal
setting.

Finally, typology gives no dependable handle. The intuitive story, that isolated
languages want fewer partners, is not what we see; if anything the two most
isolated languages want \emph{more}, but that rests on two data points and
disappears in the full set. What comes closest to tracking a language's optimum
is how well it is served on its own rather than its typology, but at ten
languages even that is a direction and not a rule ($r=-0.58$, $p=0.08$).

\section{Conclusion}

Lexical normalization has a curse of multilinguality. Training a single
fixed-size model on more languages helps only briefly: per-language accuracy
peaks at a handful of languages and then falls. A control that holds the data
fixed shows the fall is not a matter of data volume but of many languages
sharing one fixed model. A language's typological distance from the rest does
not reliably tell you where its sweet spot lies. For anyone building a compact
multilingual normalizer, the takeaway is to use fewer languages rather than pool
everything into one model.

\section*{Limitations}

Our study fixes a single small model size; the peak we observe is expected to
shift with capacity, and we do not map that dependence here. Relatedly, our
design does not separate the mechanisms behind the decline. Fewer effective parameters per language, less training exposure per language under a fixed budget, and interference between languages all move together as languages are added, so we treat them jointly as capacity rather than pin down which
dominates. We use one model family, a character-level Transformer trained from scratch, and
results may differ for large pretrained byte-level models. On two
of the twelve languages
(Spanish, Italian) our small model underperforms the leave-as-is baseline; we
report them transparently but exclude them from the aggregate, and a larger or
pretrained model might well normalize them successfully. During the sweep we
estimate test ERR on a fixed sample of tokens per language for efficiency, and we
average over random language subsets rather than all possible subsets. That
averaging is deliberate, since our question is how many partners a language
wants, but it does mean we cannot say \emph{which} partners help: the design
marginalizes over subset composition by construction, and we did not retain
per-subset records that would let us recover it. Identifying good and bad
language pairings is a natural follow-up and would need the sweep re-run with
subset-level logging.
Finally, the typology analysis covers only twelve languages, mostly
European, uses syntactic \textsc{URIEL} vectors, and maps the two code-switched
datasets onto a single matrix language; with so few and so clustered a sample,
we treat the typology question as open rather than settled.

\bibliographystyle{acl_natbib}
\bibliography{references}

\end{document}